\documentclass[10pt,twocolumn,letterpaper]{article}
\usepackage{cvpr}
\usepackage{times}
\usepackage{epsfig}
\usepackage{graphicx}
\usepackage{amsmath}
\usepackage{amssymb}
\usepackage{multirow}
\usepackage{bm}
\usepackage{enumerate}
\usepackage{amsmath}
\usepackage{mathrsfs}
\usepackage{algorithm}
\usepackage{algpseudocode}
\usepackage{graphics}
\usepackage{color}
\usepackage{colortbl}
\usepackage{soul}
\usepackage{booktabs}
\definecolor{mygray}{gray}{.9}

\usepackage[pagebackref=true,breaklinks=true,letterpaper=true,colorlinks,bookmarks=false]{hyperref}

\cvprfinalcopy 

\def\cvprPaperID{1474} 

\ifcvprfinal\fi
\begin{document}

\title{Data-Free Knowledge Amalgamation via Group-Stack Dual-GAN}

\author{Jingwen Ye\textsuperscript{1}, Yixin Ji\textsuperscript{1}, Xinchao Wang\textsuperscript{2}, Xin Gao\textsuperscript{3}, Mingli Song\textsuperscript{1}\\
\textsuperscript{1}College of Computer Science and Technology, Zhejiang University, Hangzhou, China\\
\textsuperscript{2}Department of Computer Science, Stevens Institute of Technology, New Jersey, United States \\
\textsuperscript{3}Alibaba Group, Hangzhou, China\\
\{yejingwen, jiyixin, brooksong\}@zju.edu.cn, xinchao.w@gmail.com, zimu.gx@alibaba-inc.com}


\maketitle
\thispagestyle{empty}

\begin{abstract}


Recent advances in deep learning have provided procedures for learning one network to amalgamate multiple streams of knowledge from the pre-trained Convolutional Neural Network (CNN) models, thus reduce the annotation cost. However, almost all existing methods demand massive training data, which may be unavailable due to privacy or transmission issues. In this paper, we propose a data-free knowledge amalgamate strategy to craft a well-behaved multi-task student network from multiple single/multi-task teachers. The main idea is to construct the group-stack generative adversarial networks (GANs) which have two dual generators. First one generator is trained to collect the knowledge by reconstructing the images approximating the original dataset utilized for pre-training the teachers. Then a dual generator is trained by taking the output from the former generator as input. Finally we treat the dual part generator as the target network and regroup it.
As demonstrated on several benchmarks of multi-label classification, the proposed method without any training data achieves the surprisingly competitive results, even compared with some full-supervised methods.

\end{abstract}

\section{Introduction}
In the past few years, deep convolutional neural networks (CNNs) have been widely used to achieve state-of-the-art performances in various artificial intelligent applications, such as tracking~\cite{Nam2015LearningMC,bertinetto2016fully}, classification~\cite{krizhevsky2012imagenet} and segmentation~\cite{ye2018finer-net:,Ye2019EdgeSensitiveHC}.
The success of the widely used CNNs, however, heavily relies on heavy computation and storage as well as a massive number of human annotations, sometimes even up to the scales of tens of millions such as those of ImageNet. However, in many cases of the real-world applications, the training data or annotations are confidential and therefore, unavailable to the public.

\begin{figure}[t]
\centering
\includegraphics[scale = 0.42]{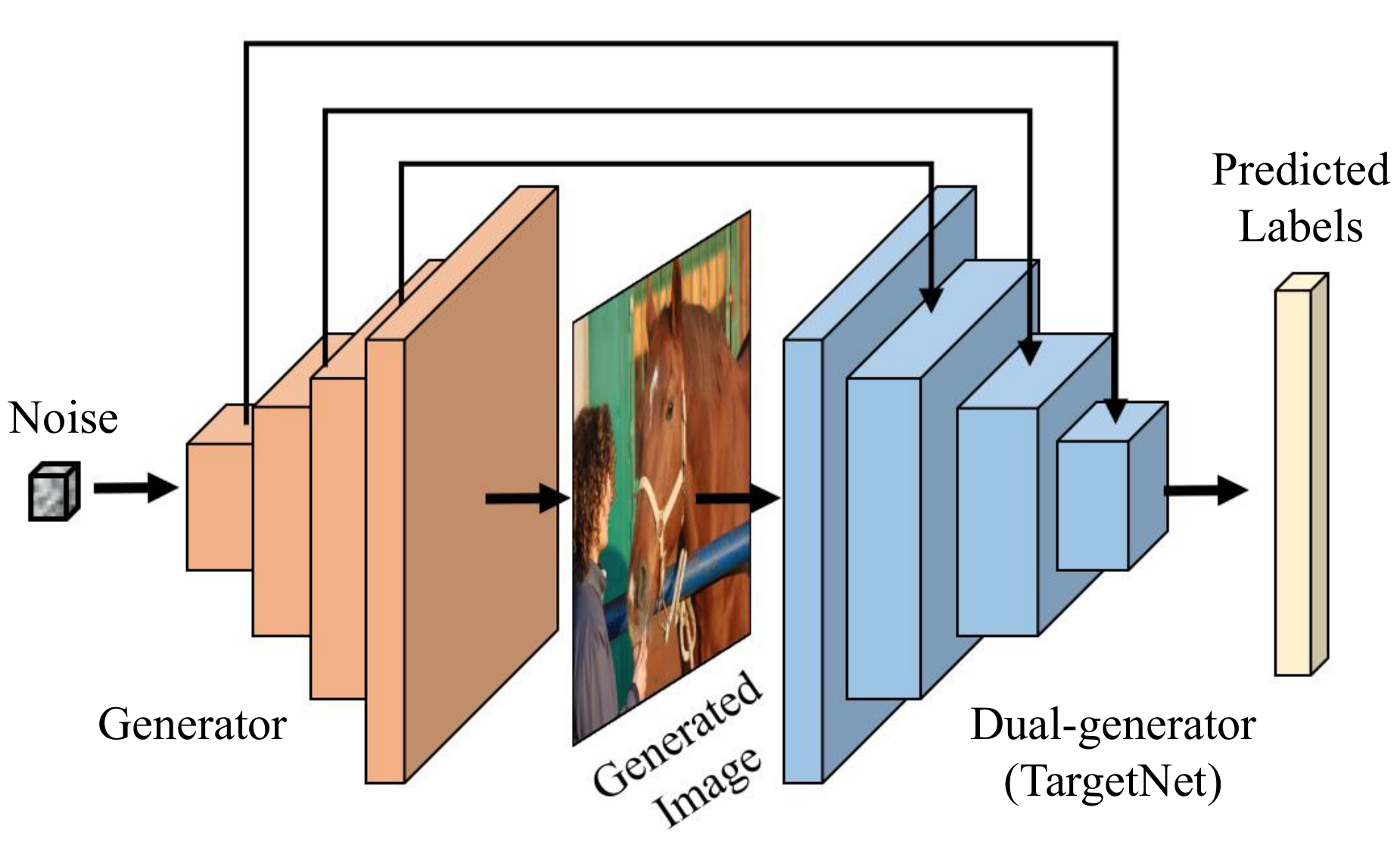}
\caption{The architecture of Dual-GAN where the corresponding discriminator is not depicted. The generator (left) synthesizes the images, while the dual-generator (right) produces the label vectors.}
\label{fig:KA}
\end{figure}

To alleviate the re-training and reproducing effort, various effective approaches have been proposed recently. For example, the usual transfer learning approach~\cite{Yosinski2014HowTA,wen2019exploiting} utilizes the pre-trained base network and then copies its first few layers to those of a target network to train toward the target task. Besides, the seminal work of knowledge distillation (KD)~\cite{hinton2015distilling} learns a compact student model by the aid of the soft labels obtained from the teachers. Other than these works that learn from one single network, knowledge amalgamation (KA)~\cite{Ye_Amalgamating_2019} is proposed to effectively reuse multiple pre-trained networks, and make the learned multi-knowledge settle well in one single network. Besides, the process of KA doesn't require any annotations, where only the intermediate features are treated as knowledge to guide training.


Training with KA demands for the unlabeled dataset only, which reduces the annotation cost largely. But it is also a common situation with no access to any training samples, even the related raw input, due to the extreme privacy policy or other irresistible factors. For example, building the face detection system, it is a violation of portraits right to publish the users' profile photos. And taking the unrelated dataset as replacement leads to unsatisfying results for the existence of the domain gap. So, in this paper, we investigate a practical strategy to train a customized network with the knowledge amalgamated from the teachers, where neither the annotations nor the input images are required.

In the field of training without data, only a few researches have been carried out, most of which work on the network compression~\cite{Nagel2019DataFreeQT,lopes2017data-free}. For example, DAFL~\cite{DAFL} applies a modified GAN to generate the input images and the corresponding annotations. Nevertheless, these methods do work on the simple datasets like MNIST~\cite{lecun1998gradient} and CIFAR-10~\cite{krizhevsky2009learning}, but go under more complicated datasets with more difficult tasks. In this paper, we propose a new data-free knowledge amalgamation framework for training the target network, which is realized by the dual-GAN in Fig.~\ref{fig:KA}. Firstly, we train a generator to amalgamate the knowledge from the pre-trained teachers. Secondly, the generator transfers the learned knowledge to the dual-generator (TargetNet) by the generated samples. Finally, the final target network is extracted from the dual-generator.

To this end, we bring forward a feasible data-free framework into knowledge amalgamation. Our contribution is therefore an effective approach to training a student model termed TargetNet, without human annotations, even without any real input data, that amalgamates the knowledge from a pool of teachers working on different tasks. The procedure is to first collect the amalgamated knowledge into the GAN and then pass it through to TargetNet. The constructed GAN is designed into the dual-architecture consisting of several unique groups, with the intermediate features generation improving the reliability.

\section{Related Work}
In this section, we briefly review the recent approaches in multi-label learning, the knowledge-based methods and several data-free methods.

\textbf{Multi-task Learning.}
Multi-task learning(MTL) has been widely studied in many fields, including computer vision~\cite{Gong2014Efficient,argyriou2008convex},
natural language processing~\cite{collobert2008a}, and machine learning~\cite{tian2015pedestrian,zhang2014facial}. The main difficulty in MTL is how to well describe the hierarchical relations among tasks and effectively learn model parameters under the regularization frameworks.

One reliable way is to utilize the tree structure. For example, Zhang et al.~\cite{Han2014Encoding} proposes a probabilistic tree sparsity model that utilizes the tree structure to obtain the sparse solution. More recently, ML-forest~\cite{wu2016ml-forest:} is proposed to learn an ensemble of hierarchical multi-label classifier trees to reveal the intrinsic label dependencies.

Another popular way has been focused on fusing MTL with CNN to learn the shared features and the task-specific models.
For example, Zhang et al.~\cite{zhang2014facial} proposes a deep CNN for joint face detection, pose estimation, and landmark localization. Misra et al.~\cite{misra2016cross-stitch} propose a cross-stitch network for MTL to learn an optimal combination of shared and task-specific representations. In~\cite{zhang2014improving}, a task-constrained deep network is developed for landmark detection with facial attribute classifications as the side tasks. Zhao et al.~\cite{Zhao2018A} proposes a multi-task learning system to jointly train the task of image captioning and two other related auxiliary tasks which help to enhance the CNN encoder and the RNN decoder in the image captioning model. With the consideration of a large number of possible label sets, most multi-task learning methods require sufficient labeled training samples. Multi-label co-training~\cite{yuying2018multi} introduces a semi-supervised method that leverages information concerning the co-occurrence of pairwise labels. To reduce the annotation cost, Durand et al.~\cite{Durand2019LearningAD} propose to train a model with a partial label with a new classification loss that exploits the proportion of known labels per example.

\textbf{Knowledge-based Learning.}
First proposed in~\cite{hinton2015distilling}, knowledge distillation aims at training a student model of a compact size
by learning from a larger teacher model or a set of teachers handling the same task, and thus finds its important application in deep model compression~\cite{Yu2017OnCD}.
More recently, the work of~\cite{gao2017knowledge} introduces a multi-teacher and single-student knowledge concentration approach. The work of~\cite{shen2019amalgamating}, on the other hand, trains a student classifier by learning from multiple teachers with different classes.

In order to handle the multi-task problem in one single network, the work of~\cite{ye2019student} proposes an effective method to train the student network on multiple scene understanding tasks, which leads to better performance than the teachers. To make it further, Ye et al.~\cite{Ye_Amalgamating_2019} apply a two-step filter strategy to customize the arbitrary task set on TargetNet.

\begin{figure*}[t]
\centering
\includegraphics[scale = 0.5]{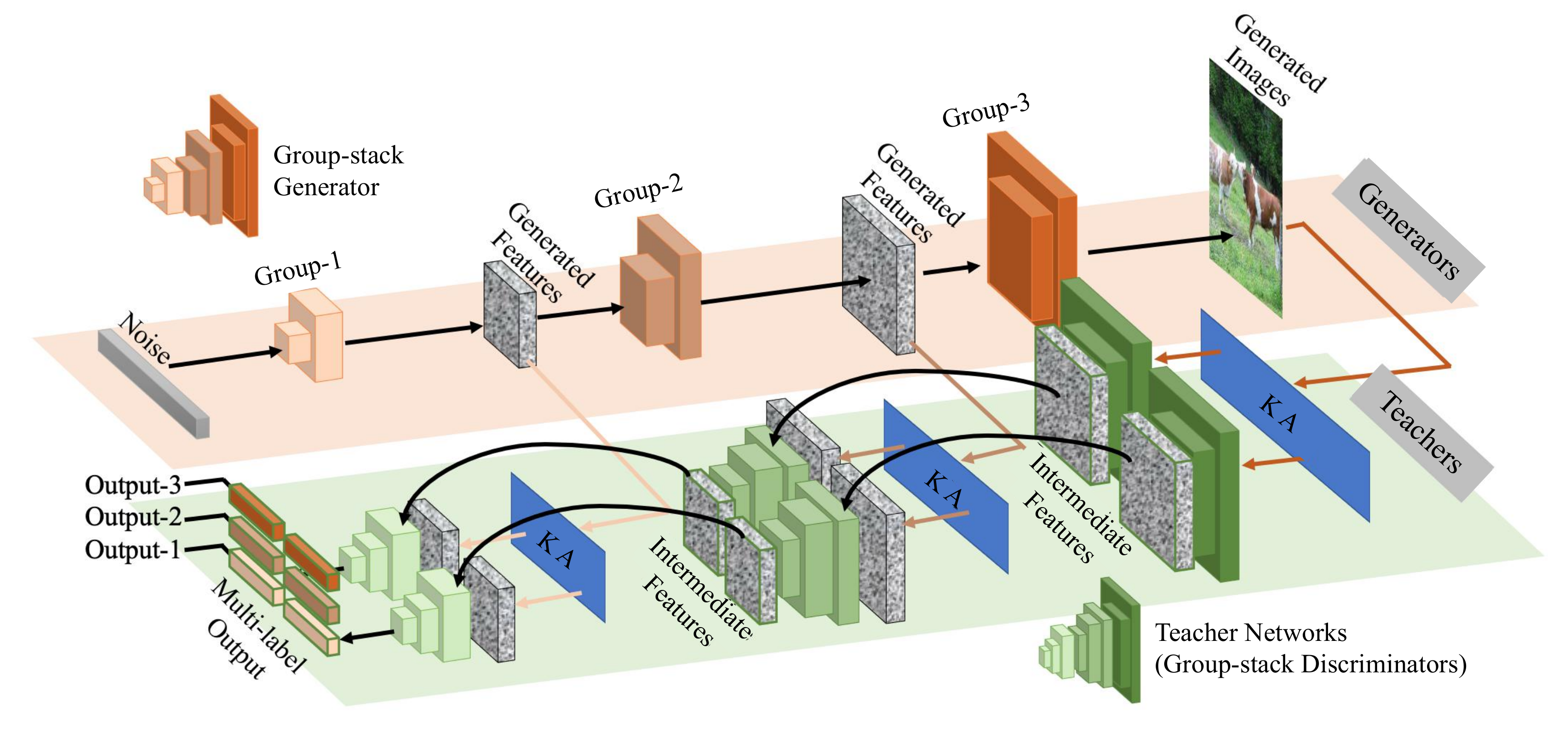}
\caption{Training the group-stack GAN with the knowledge amalgamated from the teachers. The figure exhibits the situation that a three-group GAN is trained with two teachers as corresponding multiple group-stack discriminators.}
\label{fig:groupgan}
\end{figure*}

\textbf{Data-free Learning.}
For the strict premise of leaving out the whole dataset, there are only a few methods proposed for training deep neural networks without the original training dataset. For example, Srinivas et al.~\cite{Srinivas2015DatafreePP} choose to minimize the expected squared difference of logits, which makes it possible to avoid using any training data for model compression. In addition, Lopes et al.~\cite{lopes2017data-free} attempt to regenerate batches of data based on meta data collected at training time describing the activation of the network, rather than relying on the original dataset to guide the optimization.

In addition, utilizing generative models is thought to be a more feasible way with regard to the data-free demand. GAN~\cite{Goodfellow2014GenerativeAN} has shown the capability of generating better high-quality images~\cite{wang2016generative,karras2018progressive}, compared to existing methods such as Restricted Boltzmann Machines~\cite{hinton2006a} and Deep Boltzmann Machines~\cite{Salakhutdinov2012ABW}. A vanilla GAN model~\cite{Goodfellow2014GenerativeAN} has two important components, i.e. a generator and a discriminator.
The goal of generator is to generate photo-realistic images from
a noise vector, while the corresponding discriminator is trying to distinguish between a real image and the image generated by the generator. DAFL~\cite{DAFL} utilizes GAN in the scheme of teacher-student learning, which treats the original teacher network as the discriminator and generates the training samples for the light-weight student.


\section{Problem Definition}

In this work, we aim to explore a more effective approach to train the student network (TargetNet), without any annotations, even without any training input images, only utilizing the knowledge amalgamated from the pre-trained teachers. The TargetNet is designed to deal with multiple tasks, and particularly, we focus on the multi-label classification problem, tending to learn a customized multi-branch network that can recognize all labels selected from separate teachers.

\textbf{Notations.}
We denote by $C$ the number of the customized categories, and $Y_{cst}=\{y_1,y_2,...,y_C\}\subseteq \{0,1\}^C$ as the label vector. The problem we address here is to train the TargetNet $\mathcal{T}$ which can simultaneously handle multiple tasks on the customized label set $Y_{cst}$. TargetNet amalgamates the knowledge from $M$ pre-trained teachers, which are denoted as $\mathcal{A}=\{\mathcal{A}_1,\mathcal{A}_2,...,\mathcal{A}_M\}$. For each teacher $m$, a $T_m$-label classification task $Y_m=\{y_m^1,y_m^2,...,y_m^{T_m}\}$ is learned in advance. Then, the customized label set in TargetNet and those in the teacher networks are in the constraint: $Y_{cst}\subseteq\bigcup_{m=1}^MY_m$, which reveals that either the full or the subset of classification labels is alternative for making up the customized task set.

Specifically, we use $F_m^b$ to denote the feature maps in the $b$-th block of the $m$-th pre-trained teacher, which are the knowledge to be amalgamated for the student's training.

\section{Proposed Method}
In this section, we will describe the details of the proposed data-free framework for training a target network with the knowledge amalgamated from the pre-trained teachers, which is realized by constructing the group-stack GAN in the dual architecture.

The process of obtaining the well-behaved TargetNet with the proposed data-free framework contains three steps. In the first step, we train the generator $G$ with knowledge amalgamation in the adversarial way, where the images in the same distribution of the original dataset can be manufactured. In the second step, the dual generator $\mathcal{T}$ is trained with the generated samples from $G$ in the block-wise way to produce multiple predict labels. This dual architecture containing two sub-generators can be denoted as:
\begin{equation}
\begin{split}
     G(z)&:z\rightarrow \mathcal{I}\\
     \mathcal{T}(\mathcal{I})&:\mathcal{I}\rightarrow Y_{cst},
\end{split}
\end{equation}
where $\mathcal{I}$ donates the image, $z$ is the random noise and $Y_{cst}$ is the predicted labels.

Finally in the third step, after training the whole dual-GAN, we take and modify the dual-generator as TargetNet for classifying the customized label set $Y_{cst}$. In this way, the specific GAN is embedded into the knowledge amalgamation training, thus makes it unconstrained for the training data.

\subsection{Amalgamating GAN}
In order to learn the customized TargetNet while avoiding using the real data, we choose first to amalgamate the knowledge from multiple teachers in an extra container (GAN).


Let's begin with the arbitrary vanilla GAN. The original formulation of the GAN is a minimax game between a generator, $G(z):z\rightarrow \mathcal{I}$ and a discriminator, $D(x):\mathcal{I}\rightarrow [0,1]$, and the objective function can be defined as:
\begin{equation}
\label{eq:gan}
\begin{split}
\mathcal{L}_{GAN}&= \mathbb{E}_{x\sim p_{data}(\mathcal{I})}[\log D(\mathcal{I})]\\
&+\mathbb{E}_{z\sim p_z(z)}[\log(1-D(G(z)))].
\end{split}
\end{equation}

For the absence of the real data, it is infeasible to perform the traditional training by Eq.~\ref{eq:gan}. Besides, in order to amalgamate multiple streams of knowledge into the generator, several modifications have been made as follows.

\textbf{Group-stack GAN.}
The first modification is the group-stack architecture. In this paper, the generator is designed to generate not only the images suitable for TargetNet's training, but also the intermediate activations aligned with the teachers. Thus, we set $B$ as the total group number of the generator, which is the same as the block numbers of the teacher and the student networks. In this way, the generator can be denoted as a stack of $B$ groups $\{G^1,G^2,...,G^B\}$, from which both the image $\mathcal{I}_{gan}$ and the consequent activations $F_{gan}^j$ at group $j$ are synthesised from a random noise $z$:
\begin{equation}
\label{eq:generateF}
\begin{split}
F_{gan}^{1}&=G^1(z)\\
F_{gan}^{j}&=G^j(F_{gan}^{j-1})\quad 1<j\le B,
\end{split}
\end{equation}
when $j=B$, the output of the $B$-th group $G^B$ is $F_{gan}^B$, which is also thought as the final generated image $\mathcal{I}_{gan}$.

Since the generator is in the group-stack architecture, the symmetric discriminator consisting of several groups is built. For each group $G^j$, the corresponding discriminator is $D_j$, and the group adversarial pair is presented as  $[\{G^1,D^1\},\{G^2,D^2\},...,\{G^B,D^B\}]$. In this way, the satisfying $G^{j*}$ can be acquired by:
\begin{equation}
G^{j*}=\arg\min_{G^j}\mathbb{E}_{z\sim p_{z(z)}}[\log(1-D^{j*}(G^j(F_{gan}^{j-1})],
\end{equation}
where $1\le j\le B$ and $D^{j*}$ is the optimal $j$-group discriminator. Recall that we don't have the real data, training the discriminator in the normal adversarial way becomes a virtual impossibility. Therefore, we transfer it to designing a plausible loss function to calculate the difference between the generated samples and the real ones. Thus, we take the off-the-shelf teacher network $\mathcal{A}$ to constitute each $D^{j}$:
\begin{equation}
\label{eq:discriminator}
    D^{j}\leftarrow \bigcup_{i=1}^j\{\mathcal{A}^{B-j+i}\}.
\end{equation}

During the training for the group pair $\{G^j,D^j\}$, only $G^j$ is optimized with discriminator $D^j$ fixed, whose output is for classifying multiple labels. Motivated by the work of~\cite{DAFL}, we make use of several losses to constraint the output of $D^j$ to motivate the real data's response.

The raw output for $D$ is $\mathcal{O}(F_{gan})=\{y^1,y^2,...,y^C\}$, with the predict label as $t^i$:
\begin{equation}
    t^i=
    \begin{cases}
    1 & y^i \geq \epsilon \\
    0 & y^i < \epsilon
    \end{cases},
\end{equation}
where $1\le i\le C$ and $\epsilon$ is set to be $0.5$ in the experiment. Then the one-hot loss function can be defined as:
\begin{equation}
    \mathcal{L}_{oh}=\frac{1}{C}\sum_i\ell(y^i,t^i),
\end{equation}
where $\ell$ is the cross entropy loss for each label's classification. And $\mathcal{L}_{oh}$ enforces the outputs of the generated samples to be close to one-hot vectors.

In addition, the outputs need to be sparse, since an image in the real world can't be tagged with dense labels which are the descriptions for different situations. So we propose an extra discrete loss function $\mathcal{L}_{dis}$:
\begin{equation}
\mathcal{L}_{dis}=-\frac{1}{C}\sum_i|y^i|,
\end{equation}
which is also known as the L1-norm loss function.

Finally, combining all the losses, the final objective function can be obtained:
\begin{equation}
\label{eq:allgan}
\mathcal{L}_{gan}=\mathcal{L}_{oh}+\alpha\mathcal{L}_a+\beta\mathcal{L}_{ie}+\gamma\mathcal{L}_{dis},
\end{equation}
where $\alpha$, $\beta$, and $\gamma$ are the hyper parameters for balancing different loss items. $\mathcal{L}_a$ and $\mathcal{L}_{ie}$ are the activation loss function and information entropy loss function, respectively. Those losses are proposed by ~\cite{DAFL} and will be detailed introduced in the supplementary.

\textbf{Multiple Targets.}
In this paper, the TargetNet is customized to perform multi-label classifications learning from multiple teachers. So, the generator should generate samples containing multiple targets that are learned from multiple teachers. As a result, for the $j$-th group generator $G^j$, we construct multiple group-stack discriminators $\{D_1^j,D_2^j,...,D_M^j\}$ in concert with teachers specializing in different task sets by Eq.~\ref{eq:discriminator}.

In order to amalgamate multi-knowledge into the generator by $\{D_m^j\}_{m=1}^M$, the teacher-level filtering is applied to $F_{gan}^j$. And this filtering is conducted as:
\begin{equation}
\label{eq:teacherfilter}
    F_{gan}^{j,m}=f_m^j(F_{gan}^j),
\end{equation}
where the filtering function $f_m^j$ is realized by a light learnable module consisting of a global pooling layer and two fully connected layers. $F_{gan}^{j,m}$ is the filtered generated features that approaches the output feature distribution of $\mathcal{A}_m^{B-j}$, also known as the $(B-j)$-th block of $\mathcal{A}_m$. And the validity of the generated $F_{gan}^{j,m}$ is justified by $D_m^j$, whose outputs are donated as $\mathcal{O}_m(F_{gan}^{j})=D_m^j(F_{gan}^{j,m})$.

Then for the generated features $F_{gan}^j$ from $G^j$, we collect from the multi-discriminator the $M$ prediction sets $\{\mathcal{O}_1(F_{gan}^j),\mathcal{O}_2(F_{gan}^j),...,\mathcal{O}_M(F_{gan}^j)\}$, which are:
\begin{equation}
\label{eq:finalo}
\mathcal{O}_{gan}(F_{gan}^j)=\bigcup_{m=1}^M\mathcal{O}_m(F_{gan}^j),
\end{equation}
which is treated as new input to the loss Eq.~\ref{eq:allgan}, then $\mathcal{L}_{gan}^j$ is the adversarial loss for each $G^j$. Despite the fact that the generated features should appear like the ones extracted from the real data, they should also lead to the same predictions from the same input $z$. Thus, the stack-generator $\{G^1,G^2,...,G^B\}$ can be jointly optimized by the final loss:
\begin{equation}
\label{eq:joint}
    \mathcal{L}_{joint}= \mathcal{L}_{gan}^B+\frac{1}{B-1}\sum_{j=1}^{B-1}\ell(\mathcal{O}_{gan}(F_{gan}^j),\mathcal{O}_{gan}(\mathcal{I}_{gan})),
\end{equation}
where the adversarial loss $\mathcal{L}_{gan}$ only calculates from the last group $\{G^B,D^B\}$. The rest part of the final loss is the cross-entropy loss that restrains the intermediate features generated from $G^1$ to $G^{B-1}$ to make the same predictions as $\mathcal{I}_{gan}$, which offsets the adversarial loss $\{\mathcal{L}_{gan}^1,...,\mathcal{L}_{gan}^{B-1}\}$.

By minimizing $\mathcal{L}_{joint}$, the optimal generator $G$ can synthesis the images that have the similar activations as the real data fed to the teacher. 

\subsection{Dual-generator Training}
\label{sec:ka}
After amalgamating the multi-stream knowledge into the group-stack GAN, a set of generated training samples are obtained, which are in the form of $\mathcal{R}=\{F_{gan}^1,F_{gan}^2...,F_{gan}^{B-1}\}\cup\{\mathcal{I}_{gan}\}$ including both the generated intermediate features and the RGB image. Then the next step is to train the dual-generator $\mathcal{T}$. Also, we construct the group-discriminator for each $\mathcal{T}^b$ as:
\begin{equation}
\label{eq:dualdis}
D_{dual}^{b,m}\leftarrow \bigcup_{i=1}^{B-b}\{\mathcal{A}_m^{b+i}\},
\end{equation}
where the dual generator $\mathcal{T}$ is trained to synthesis the samples that $D_{dual}$ can't distinguished from $\mathcal{I}_{gan}$ produced by the trained $G$.


Motivated by the work of ~\cite{Ye_Amalgamating_2019}, we use the block-wise training strategy to transfer as much knowledge as possible from the generator $G$ into the dual-generator $\mathcal{T}$. That is, we divide dual-generator into $B$ blocks as $\{\mathcal{T}^1,\mathcal{T}^2,...,\mathcal{T}^B\}$, and during the training process of $\mathcal{T}^b$ $(1<b\le B)$, the whole generator $G$ and the blocks of the dual-generator $\mathcal{T}$ from $1$ to $b-1$ keep fixed.
\begin{figure}[t]
\centering
\includegraphics[scale = 0.57]{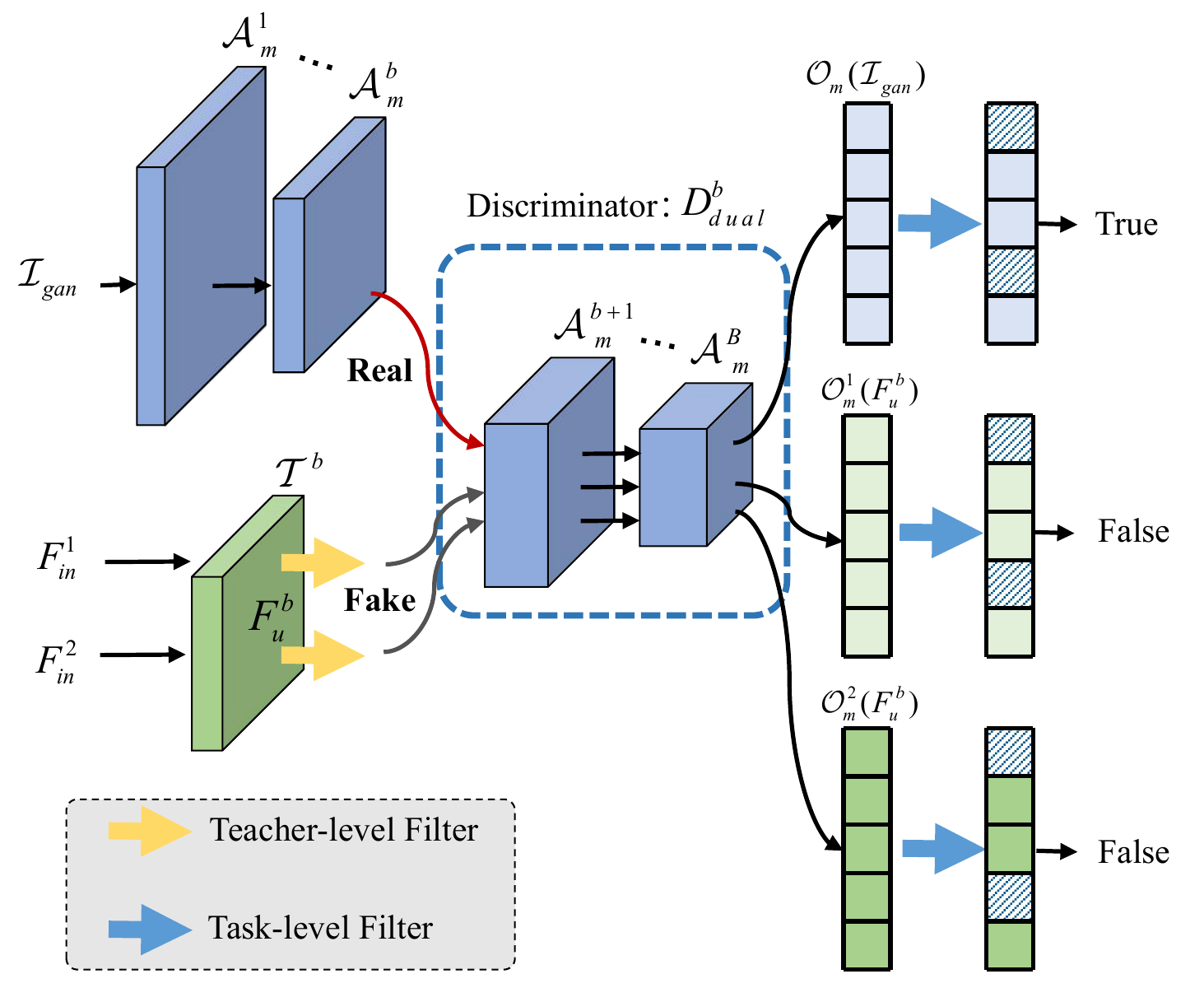}
\caption{The adversarial mode for $b$-th block of the dual-GAN. Two levels of filtering are performed to the features generated by the $b$-th group of the dual-generator $\mathcal{T}^b$.}
\label{fig:block}
\end{figure}

Take training $\mathcal{T}^b$ for example, the problem is to learn the features $F_u^b$ generated by the $b$-th block of the dual-generator. As shown in Fig.~\ref{fig:block}, which shows the adversarial for dual-GAN block $\{\mathcal{T}^b,D^b\}$, the generated $F_u^b$ is treated as `fake' to the discriminator with $\mathcal{I}_{gan}$ as the `real' data in this dual part GAN. To do this, a two-level filtering method is applied to $F_u^b$. The first step is the teacher-level filtering for multi-target demand, which transforms $F_u^b$ to $m$ teacher streams $F_{u}^{b,m}$ by $F_{u}^{b,m}=f_m^b(F_u^b)$, as defined in Eq.~\ref{eq:teacherfilter}. The second level is the task-level filtering conducted after the last few fully connected layers of the corresponding discriminator, which is established for the constraint of $Y_{cst}\subseteq\bigcup_{m=1}^MY_m$. We feed the generated features $F_u^b$ to the corresponding discriminator $D_{dual}^{b,m}$, and derive the predictions
$\mathcal{O}_m(F_{u}^{b})$. Then the task-level filtering $g_m$ is applied to meet the task customizable demand and embedded into the $m$-th branch block-wise adversarial loss:
\begin{equation}
\mathcal{L}_{dual}^{b,m}= \ell (g_m(\mathcal{O}_m(F_{u}^{b})),g_m(\mathcal{O}_m(\mathcal{I}_{gan}))),
\end{equation}
where $g_m$ is utilized to select the needed predictions for customized tasks in the constraint: $Y_{cst}=\bigcup_{m=1}^Mg_m(Y_m)$. So the block-wise loss for updating dual-generator $\mathcal{T}^b$ from multiple branches can be denoted as:
\begin{equation}
\label{eq:kaloss}
\mathcal{L}_{dual}^b=\sum_m\lambda_m\cdot\mathcal{L}_{dual}^{b,m},
\end{equation}
where we set $\lambda_m=1$ for $m\in\{1,...,M\}$.

Since the existence of the links between the generators $G$ and $\mathcal{T}$, the input to the present block has two streams:
\begin{equation}
\label{eq:Fin}
\begin{split}
    F_{in}^1&=\mathcal{T}^{b-1}\mathcal{T}^{...}\mathcal{T}^{1}(\mathcal{I}_{gan}),\\
    F_{in}^2&=G^{B+1-b}G^{...}G^{1}(z),
\end{split}
\end{equation}
where $F_{in}^1$ is obtained from $\mathcal{I}_{gan}$ and $F_{in}^2$ is generated by $G^{B+1-b}$ that indicates the links in Fig.~\ref{fig:KA}. Then according to the different inputs to $\mathcal{T}^b$, the final loss can be rewritten from Eq.~\ref{eq:kaloss} as:
\begin{equation}
\label{eq:final}
\mathcal{L}_u^b=\lambda_{in}^1\mathcal{L}_{dual}^{b,m}(F_{in}^1)+\lambda_{in}^2\mathcal{L}_{dual}^{b,m}(F_{in}^2),
\end{equation}
where $\lambda_{in}^1$ and $\lambda_{in}^2$ are the balancing weights. With Eq.~\ref{eq:final}, parameter updating takes place within block $b$ and the corresponding learnable teacher-level filters $\{f_1^b,...,f_M^b\}$, and blocks from 1 to $b-1$ remain unchanged.

\subsection{TargetNet Regrouping}

Once the training of all the $B$ blocks from the dual-generator has been completed by Eq.~\ref{eq:kaloss}, the whole GAN in the dual architecture is obtained. Then we can acquire the embryo TargetNet from the dual-part of the generator $\{\mathcal{T}^1,\mathcal{T}^2,...,\mathcal{T}^B\}$ as well as a series of loss convergence values $\{\eta^{1,m},\eta^{2,m},...,\eta^{B,m}\}$ calculated from $\mathcal{L}_{dual}^{b,m}$ at each block for teacher $\mathcal{A}_m$. $\mathcal{T}_u$ treats all the sets of labels equal and produces the final predictions at the last block $\mathcal{T}_u^B$, which is against our expectation to make them hierarchical. Then we do the branching out on $\mathcal{T}$ and the branch-out blocks are taken to be:
\begin{equation}
\label{eq:branchout}
\begin{split}
S_m=\arg\min_{b\in[1,B]}\eta^{b,m}.
\end{split}
\end{equation}

Once the branch-out spot $S_m$ for task set $g_m(Y_m)$ is determined, we regroup the branched task-specific TargetNet and keep the corresponding teacher-wise filter $f_m^{S_m}$ for connecting the TargetNet and the corresponding teacher $\mathcal{A}_m$ as:
\begin{equation}
\label{eq:link}
\mathcal{T}_m=[\{\mathcal{T}^1,...,\mathcal{T}^{S_m}\},f_m^{S_m},\{\mathcal{A}_m^{S_m+1},...,\mathcal{A}_m^B\}],
\end{equation}
where $\mathcal{T}_m$ is specialized for the task set $g_m(Y_m)$.

Finally, we can regroup the final TaregtNet $\mathcal{T}_u$ in the hierarchical architecture as:
\begin{equation}
    \mathcal{T}_u=\bigcup_{m=1}^M\{\mathcal{T}_m\},
\end{equation}
which shares the first few $\min\{S_1,S_2,...S_M\}$ blocks and branches out in the following blocks for different tasks.

\begin{algorithm}[t]
  \caption{Data-free Knowledge Amalgamation}
  \label{alg::training}
  \begin{algorithmic}[1]
    \Require
      $\{\mathcal{A}_1,\mathcal{A}_2,...,\mathcal{A}_M\}$: $M$ pre-trained teacher set;
      $\{\mathcal{A}_m^1,\mathcal{A}_m^2,...,\mathcal{A}_m^B\}$: $m$-th teacher divided into $B$ blocks for classifying $Y_m$;
      $Y_{cst}$: customized label set.
    \State Initial the group-stack generator $\{G^1,G^2,...,G^B\}$ and the dual generator $\{\mathcal{T}^1,\mathcal{T}^2,...,\mathcal{T}^B\}$;
    \State \emph{\textbf{Step I: Training the Generator.}}
    \State Randomly sample the noise $z$ and initial $F_{gan}^0\leftarrow z$;
    \For{group $j=1:B$}
        \State Build $D^j$ for $G^j$ by Eq.~\ref{eq:discriminator} and
        get $F_{gan}^j$ by Eq.~\ref{eq:generateF};
        \State For all $m\in[1,M]$, get $F_{gan}^{j,m}$ by Eq.~\ref{eq:teacherfilter};
        \State Get predictions $\mathcal{O}_{gan}(F_{gan}^j)$ by Eq.~\ref{eq:finalo};
    \EndFor
    \State Calculate loss $\mathcal{L}_{joint}$ by Eq.~\ref{eq:joint} to optimize $G$;
    \State \emph{\textbf{Step II: Training the Dual-generator.}}
    \State Acquire $\{F_{gan}^1,F_{gan}^2,...F_{gan}^{B-1}\}\cup\mathcal{I}_{gan}$ from $z$;
    \State Feed $\mathcal{I}_{gan}$ into $\mathcal{A}_m$ and initial $F_u^0\leftarrow\mathcal{I}_{gan}$;
    \For{block $b=1:B$}
      \State Build $D_{dual}^b$ for $\mathcal{T}^b$ by Eq.~\ref{eq:dualdis};
      \State Obtain $F_{in}^1$, $F_{in}^2$ by Eq.~\ref{eq:Fin} and get $F_u^b=\mathcal{T}^b(F_{in})$;
      \State Do the two-level filtering to $F_u^b$;
      \State Optimize $\mathcal{T}^b$ by the loss Eq.~\ref{eq:final};
    \EndFor
    \State \emph{\textbf{Step III: Branching out.}}
    \State Collect the convergence loss value $\eta^{k,m}$.
    \State Find branch-out point $S_m$ by Eq.~\ref{eq:branchout};
    \State Get task-specific $\mathcal{T}_m$ by Eq.~\ref{eq:link};
    \State Group and fine-tune $\mathcal{T}_u=\bigcup_{m=1}^M\{\mathcal{T}_m\}$.
    \Ensure
      $\mathcal{T}_u$: task-specific hierarchical TargetNet.
  \end{algorithmic}
\end{algorithm}

The proposed data-free knowledge amalgamation method is first to amalgamate a group-stack multi-target GAN, and then utilizing the generated intermediate feature maps as well as the final output images to train the dual-generator. Finally, the TargetNet can be acquired by modifying the trained generator. The whole process of data-free training TargetNet is exhibited in Algorithm~\ref{alg::training}.

\begin{table*}[t]
\centering
\small
\caption{Comparisons of the classification results (AP in \%) on the randomly selected 10-label set with other methods on VOC 2007.}
\begin{tabular}{l|c|ccccccccccc}
\toprule
 Scenario &Required Data & plane & bike & bird& boat & bus & car & horse& motor &person & train &mAP\\
\hline\hline
TeacherNet&VOC2007&94.2&81.9&82.4&81.9&76.2&89.2&89.4&82.4&93.1&89.9&86.0\\
KA&Unlabeled VOC2007&94.3&83.6&82.5&82.0&76.7&89.2&89.5&82.5&93.2&90.1&86.4\\
\hline
Similar Data&CIFAR100&80.4&65.2&71.3&68.0&66.9&81.3&62.1&56.1&87.2&79.5&71.8\\
Diff Data&CityScape&5.4&47.8&12.8&6.1&20.0&63.4&11.1&13.5&71.0&20.4&27.2\\
Random Noise&None&1.2&3.5&34.5&13.5&7.8&9.2&23.7&29.0&3.5&19.8& 14.6\\
DAFL &None&32.8&58.0&37.3&50.5&28.9&63.1&51.8&41.6&75.1&29.4&46.8\\
\hline
Data-free KA &None&84.9&67.1&66.9&67.1&58.4&83.0&72.5&58.0&80.4&77.2&73.6\\
\bottomrule
\end{tabular}
\label{tab:vocsingle}
\end{table*}

\section{Experiments and Results}
Here we provide our experimental settings and results. More results can be found in our supplementary material.

\subsection{Experimental Settings}
\textbf{Datasets.}
In this paper, we evaluate the proposed method on several standard multi-label datasets: the PASCAL Visual Object Classes 2007 (VOC2007)~\cite{pascal-voc-2007} and Microsoft Common Objects in Context (MS-COCO)~\cite{Lin2014Microsoft} dataset. These two datasets contain 9,963 images and 123,287 images respectively, where there are totally 20 object concepts annotated for VOC2007 and 80 for MS-COCO. In each dataset, we use the training set for pre-training the teachers. For TargetNet, we don't refer to any real data and only draw support from the knowledge amalgamated from the teachers.


\textbf{Implementation details.}
We implemented our model using TensorFlow with a NVIDIA M6000 of 24G memory.
We adopt the poly learning rate policy.
We set the base learning rate to 0.01, the power to 0.9,
and the weight decay to $5e-3$. Due to limited physical memory on GPU cards and to assure effectiveness, we set the batch size to be 16 during training. The TargetNet which is in the same architecture as the teachers (ResNet101~\cite{he2016identity}), is initialized with parameters pre-trained on ImageNet.

\subsection{Experimental Results}

\subsubsection{Performance on VOC 2007}
\textbf{Customized Tasks of Single Teacher.}
There are a total 20 labels in Pascal VOC 2007, we pre-trained one teacher network (`Pretrained Teacher') on the total label set. Then the TargetNet is designed to classify 10 labels randomly selected from the 20 labels. Then we compare the results with the following methods:\\
\emph{[KA]}: The block-wise training method proposed by~\cite{Ye_Amalgamating_2019} based on unlabeled original dataset;\\
\emph{[Similar Data]}: The method trained with the real dataset similar with the original one;\\
\emph{[Diff Data]}: The method trained with the real dataset that is quit different from the original one;\\
\emph{[Random Noise]}: The method trained with Random Noise;\\
\emph{[DAFL]}: The data-free method proposed by~\cite{DAFL}.

Then all results are shown in Table~\ref{tab:vocsingle} for comparison, where the AP for each label and the mAP for all are depicted. `KA' outperforms the teacher by 0.4\%, which trains TargetNet with the unlabeled part of VOC 2007. And the classification results trained from the substitute datasets (`Similar Data' and `Diff Data') show that using an irrelated dataset to replace the original one is not reliable, since it heavily depends on the dataset choosing, which makes the final results uncontrollable and stochastic. Besides, compared with the methods (`Random Noise' and `DAFL') that don't use any training data, the effectiveness of the proposed method (`Data-free KA') can be easily proved.

\textbf{Multiple Teachers on Whole Label Set.}
In this setting, we divide the whole 20 labels randomly into two groups which are learned separately in two teacher networks.
To the best of our knowledge, we are the first to study the data-free customized-label classification. So we set the customized task set $Y_{cst}=\bigcup_{m=1}^M{Y_m}$, which makes the TargetNet deal with a normal multi-label task for comparison with other methods. We compare the performance of the proposed method against the following typical multi-label approaches: `HCP-2000'~\cite{wei2014CNN}, `HCP-VGG'~\cite{wei2016hcp:}, `RLSD+ft-RPN'~\cite{zhang2018multilabel} and `KA-ResNet'. And since we have not found another data-free or unsupervised method on multi-label classification, we refer to a related work `partial-BCE'~\cite{Durand2019LearningAD}.

Table~\ref{tab:vocall} presents the results on this setting. The proposed data-free method achieves the competitive results compared with other methods that rely on the training dataset more or less. Besides, the our accuracy results do surpass some of the full-supervised methods.

\begin{table}[t]
\centering
\small
\caption{Comparisons of the classification results (mAP in \%) on the whole 20-label on VOC 2007.}
\begin{tabular}{l|c|c}
\toprule
Methods & mAP & Description\\
\hline
HCP-2000 & 85.2& \multirow{3}{*}{Training with original dataset.}\\
HCP-VGG & 90.9 & \\
RLSD+ft-RPN & 88.5 & \\
\hline
partial-BCE & 90.7 & Training with 10\% annotations \\
\hline
KA-ResNet & 92.5 &Training with unlabeled dataset.\\
\hline
DAFL &54.9 & \multirow{2}{*}{Training without real data.}\\
\bf{Data-free KA}&87.2& \\
\bottomrule
\end{tabular}
\label{tab:vocall}
\end{table}

\begin{figure*}[t]
\centering
\includegraphics[scale = 0.9]{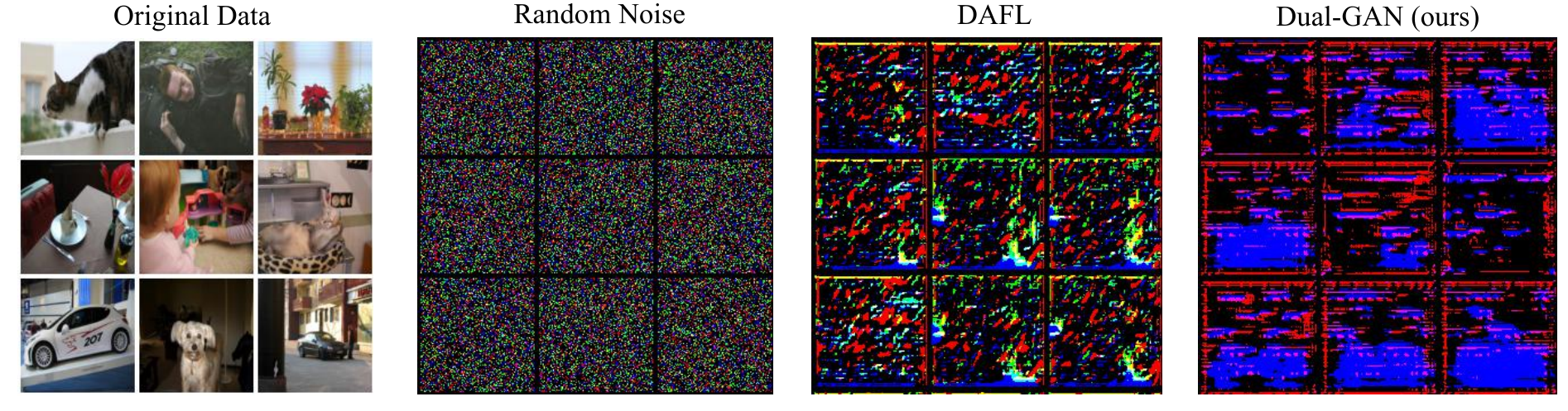}
\caption{Visualization of the generated images compared with the original dataset trained on the teachers.}
\label{fig:ganImage}
\end{figure*}

\textbf{Visualization Results.} As shown in Fig.~\ref{fig:ganImage}, we visualize the images generated from the last group of $G$, and compare them with the ones from the original data, random noise and DAFL. For lack of the real data, the trained generator is unable to synthesis the images that are thought to be `real' by human but it can synthesis the `real' ones by the convolutional networks' perception. And the images generated from `DAFL' seem to be more closer to the original data than the ones generated from the GAN proposed in this paper, but the proposed method outperforms `DAFL' a lot, which indicates the existence of the big gap between human and the neural networks.

\subsubsection{Performance on MS-COCO}

On MS-COCO dataset, we follow the work of ~\cite{wang2016cnn} to select the top $k = 3$ labels for each image.

In Table~\ref{tab:coco}, we compare the overall precision (`O-P'), recall (`O-R'), F1 (`O-F1'), and per-class precision (`C-P'), recall (`C-R'), F1 (`C-F1') with other methods. Our proposed `Data-free KA' is trained to learn from two pre-trained teachers, and outperforms some of the full-supervised methods on some metrics. Compared with the other data-free framework (`DAFL'), our method shows better results.

\begin{table}
\caption{Comparison of our method and state-of-the-art methods on the MS-COCO dataset}
\centering
\small
\begin{tabular}{l|cccccc}
\toprule
Method  & C-P & C-R & C-F1 & O-P&O-R&O-F1 \\
\midrule
RLSD+ft-RPN &67.6&57.2&62.0&70.1&63.4&66.5\\
KA-ResNet&79.8&63.5& 69.4& 83.9& 65.7&75.0\\
DAFL& 44.0&23.8&34.3&48.9&28.3&36.6\\
\midrule
Data-free KA & 66.5 & 55.9&61.0 &69.8&65.4&68.2  \\
\bottomrule
\end{tabular}
\label{tab:coco}
\end{table}

\subsubsection{Ablation study}
We now analyze the components of our approach and demonstrate their impact on the final performance. All these ablation studies are performed on VOC2007 and learned from one single teacher.

\begin{figure}[t]
\centering
\small
\includegraphics[scale = 0.355]{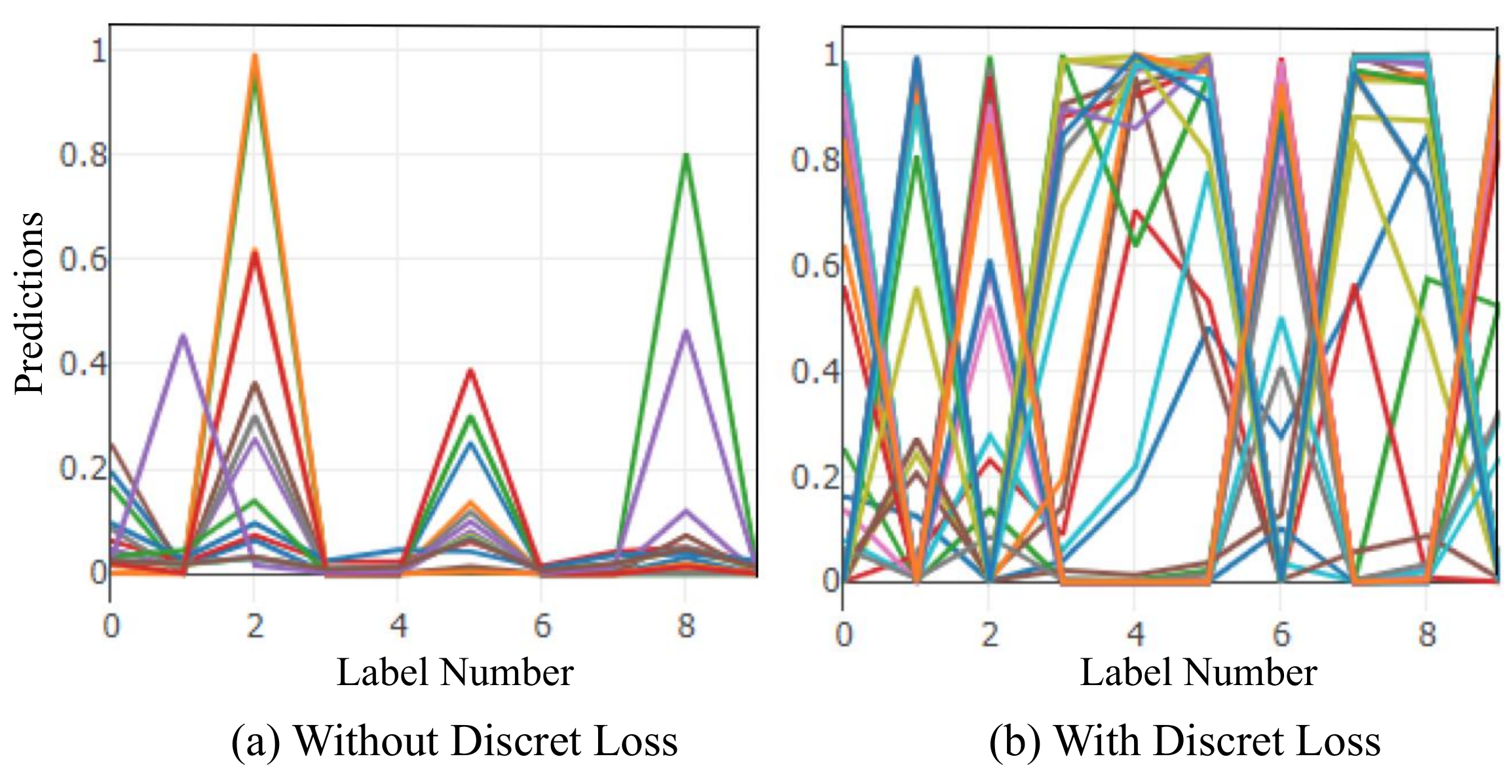}
\caption{Label distribution without/with $\mathcal{L}_{dis}$. The add of $\mathcal{L}_{dis}$ helps the label distribution of the final output more uniform.}
\label{fig:ldis}
\end{figure}

\textbf{Discrete Loss.} Compared with DAFL, we add an extra discrete loss $\mathcal{L}_{dis}$. In Fig.~\ref{fig:ldis}, the final predictions got from the teacher with the generated images are depicted. In the figure, the predictions with the discrete loss have more diversity, which means that the generated images in this way are better for the TargetNet's training.

\begin{table}[t]
\centering
\small
\caption{Ablation study of the influence on $\lambda_{in}^1$ and $\lambda_{in}^2$.}
\begin{tabular}{p{20mm}|p{15.5mm}<{\centering}p{15.5mm}<{\centering}p{15.5mm}<{\centering}}
\toprule
 $\{\lambda_{in}^1,\lambda_{in}^2\}$  & $\{1,0\}$&$\{0,1\}$&$\{1,1\}$\\
\hline
mAP (\%) & 70.2& 59.8&73.6 \\

\bottomrule
\end{tabular}
\label{tab:para}
\end{table}

\textbf{Links in Dual-GAN.} For the existence of the links between the generator and the dual-generator, the input to the dual-generator has two streams, based on which, the inputs can be divided into three cases:\\
1. $\{\lambda_{in}^1=1,\lambda_{in}^2=0\}$: The dual-generator is trained only with $F_{in}^1$ as input with no links between the generator and the dual one;\\
2. $\{\lambda_{in}^1=0,\lambda_{in}^2=1\}$: The dual-generator is trained only with $F_{in}^2$ as input with the links, but $\mathcal{I}_{gan}$ is not fed into;\\
3. $\{\lambda_{in}^1=1,\lambda_{in}^2=1\}$: The combination of 1 and 2.

As is presented in Table~\ref{tab:para}, the dual-GAN with the main link and the link between the two dual parts performs the best in the data free knowledge amalgamation.

\section{Conclusion}
In this paper, we present a data-free method to customize a network on multiple tasks, under the guidance of several pre-trained teachers. The support for training without real data is drawn from a specific GAN with two vital characteristics: one is the architecture stacked with several groups which force the generator to synthesis reliable images along with reliable intermediate activations; the other one is the dual architecture which enables the two sub generators to cooperate together effectively. Once the required GAN has been trained, the target network can be constructed immediately. Experimental results show that the proposed method is able to train a well-behaved student network while avoiding the participation of the real data.

In the future, we will explore a more effective data-free approach to train TargetNet with the knowledge directly transferred from the teachers.

\section*{Acknowledgement}
This work is supported by  National Key Research and Development Program (2018AAA0101503) , National Natural Science Foundation of China (61976186),  Key Research and Development Program of Zhejiang Province (2018C01004), and the Major Scientific Research Project of Zhejiang Lab (No. 2019KD0AC01).

\clearpage

\end{document}